\documentclass[letterpaper]{article}
\usepackage{aaai18}
\usepackage{times}
\usepackage{helvet}
\usepackage{courier}
\usepackage{url}
\usepackage{graphicx}

\usepackage{amsmath}  % math formulation
\usepackage{listings} % code extracts
\title{Modularity as a Means for Complexity Management in Neural Networks Learning}
\author{
    David Castillo-Bolado \\
    \texttt{david.castillo@siani.es}
    \And
    Cayetano Guerra-Artal \\
    \texttt{cayetano.guerra@ulpgc.es} \\
    SIANI Institute and Department of Computer Science. University of Las Palmas de Gran Canaria
    \And
    Mario Hernandez-Tejera \\
    \texttt{mario.hernandez@ulpgc.es}
}

\begin{document}

\maketitle

\begin{abstract}
Training a Neural Network (NN) with lots of parameters or intricate architectures creates undesired phenomena that complicate the optimization process. To address this issue we propose a first modular approach to NN design, wherein the NN is decomposed into a control module and several functional modules, implementing primitive operations. We illustrate the modular concept by comparing performances between a monolithic and a modular NN on a list sorting problem and show the benefits in terms of training speed, training stability and maintainability. We also discuss some questions that arise in modular NNs.
\end{abstract}

\section{Introduction}

There has been a recent boom in the development of Deep Neural Networks, promoted by the increase in computational power and parallelism and its availability to researchers. This has triggered a trend towards reaching better model performances via the growth of the number of parameters \cite{ResNet} and, in general, increment in complexity \cite{Inception} \cite{NTM}.

However, training NNs with lots of parameters emphasizes a series of undesired phenomena, such as gradient vanishing \cite{GradientVanishing}, spatial crosstalk \cite{Crosstalk} and the appearance of local minima. In addition, the more parameters a model has, the more data and computation time are required for training \cite{VCDimension}.

The research community has had notable success in coping with this scenario, often through the inclusion of priors in the network, as restrictions or conditionings. The priors are fundamental in machine learning algorithms and they have been, in fact, the main source of major breakthroughs within the field. Two well known cases are the Convolutional Neural Networks \cite{ConvNet} and the Long Short-Term Memory \cite{LSTM}. This trend has reached the extent that recent models, developed to solve problems of moderate complexity, build up on elaborated architectures that are specifically designed to the problem in question \cite{seq2seq} \cite{PixelCNN}.

But these approaches are not always sufficient and, while new techniques like attention mechanisms are now enjoying great success \cite{AttentionIsAllYouNeed}, they are integrated in monolithic approaches that tend to suffer from overspecialization. Thus Deep Neural Networks become more and more unmanageable every time they grow in complexity; impossible for modest research teams to deal with, as the state of the art is often built upon exaggerated computational resources \cite{LotsOfTeslas}.

NNs were developed by mimicking biological neural structures and functions, and have ever since continued to be inspired by brain-related research \cite{BrainInspiration}. Such neural structures are inherently modular and the human brain itself is modular in different spatial scales, as the learning process occurs in a very localized manner \cite{BrainModularity}. That is to say that the human brain is organized as functional, sparsely connected subunits. This is known to have been influenced by the impact of efficiency in evolution \cite{ModularityOrigins}.

In addition, modularity has an indispensable role in engineering and enables the building of highly complex, yet manageable systems. Modular systems can be designed, maintained and enhanced in a very controlled and methodological manner, as they ease tractability, knowledge embedding and the reuse of preexisting parts. Modules are divided according on functionality, according to the rules of high cohesion and low coupling, so new functionality will come as new modules in the system, leaving the others mostly unaltered.

In this paper we propose a way to integrate the prior of modularity into the design process of NNs. We make an initial simplified approach to modularity by working under the assumption that a problem domain can be solved based on a set of primitive operations. Using this proposal, we aim to facilitate the building of complex, yet manageable systems within the field of NNs, while enabling diverse module implementations to coexist. Such systems may evolve by allowing the exchange and addition of modules, regardless of their implementation, thus avoiding the need to always start from scratch. Our proposal should also ease the integration of knowledge in the form of handcrafted modules, or simply through the identification of primitives.

Our main contributions are:

\begin{itemize}
    \item We propose an initial approach to a general modular architecture for the design and training of complex NNs.
    \item We discuss the possibilities regarding the combination of different module implementations, maintainability and knowledge embedding.
    \item We prove that a NN designed with modularity in mind is able to train in a shorter time and is also a competitive alternative to monolithic models.
    \item We give tips and guidelines for transferring our approach to other problems.
    \item We give insights into the technical implications of modular architectures in NNs.
\end{itemize}

The code for the experiments in this paper is available at \texttt{gitlab.com/dcasbol/nn-modularization}.

\section{The modular concept}

\begin{figure}
    \centering
    \includegraphics[keepaspectratio,scale=0.7]{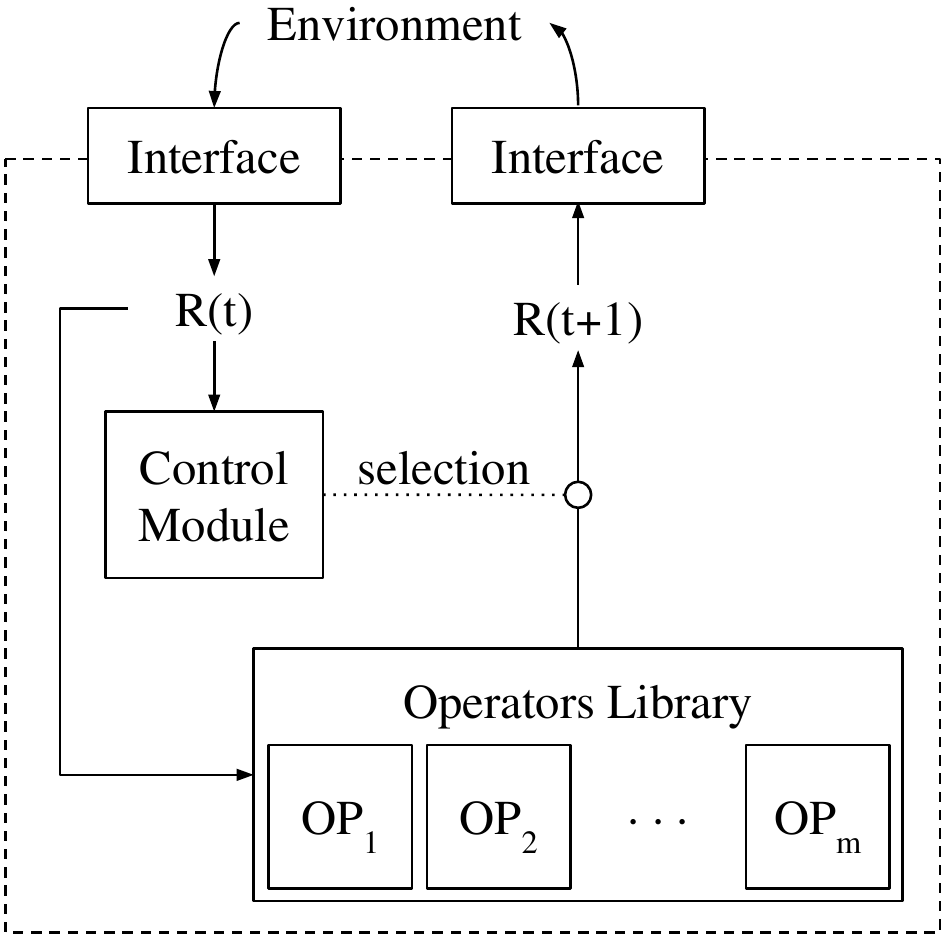}
    \caption{Perception-action loop. Each module is susceptible to being implemented by a NN. At each step, the control module selects an operator to be applied and this will generate the next environment's state.}
    \label{fig:main_loop}
\end{figure}

We are proposing a modular approach to NN design in which modularity is a key factor, as is the case in engineering projects. Our approach has many similarities to the blackboard design pattern \cite{blackboard} and is based on a perception-action loop (figure \ref{fig:main_loop}), in which the system is an agent that interacts with an environment via an interface. The environment reflects the current state of the problem, possibly including auxiliary elements such as scratchpads, markers or pointers, and the interface provides a representation $R(t)$ of it to work with. $R(t)$ is thus a sufficient representation of the environment state at time $t$. This representation will match any relevant change in the environment as soon as they occur and, if any changes were made by the agent in the representation, the interface will be able to forward them to the environment. This feedback through the environment is what closes the perception-action loop.

\begin{figure}
    \centering
    \includegraphics[keepaspectratio,width=\columnwidth]{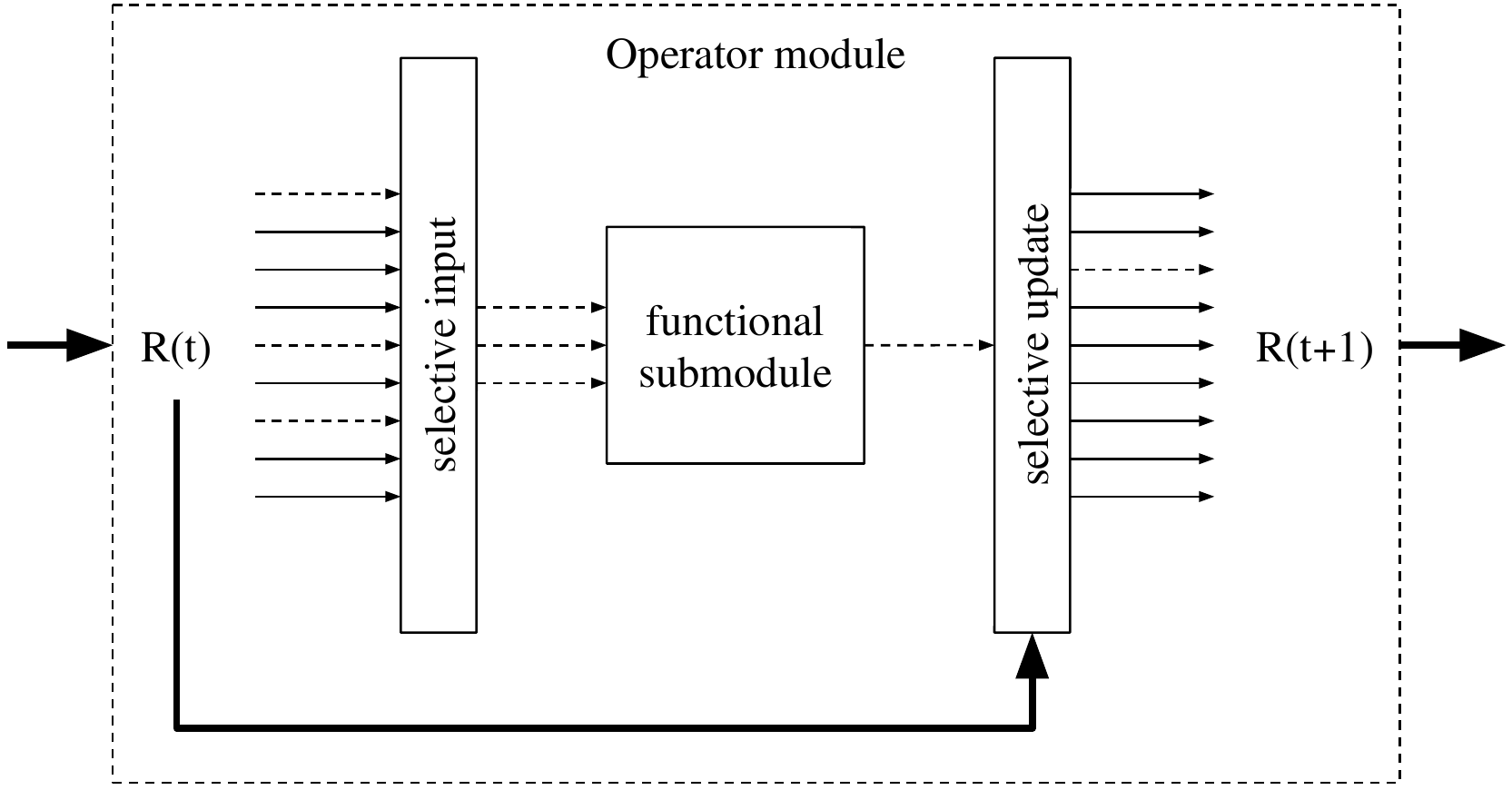}
    \caption{Detail of an operator module, composed of an input-selection submodule, a functional kernel and a selective-update submodule. Dashed lines highlight the selected data.}
    \label{fig:operator}
\end{figure}

In the middle of this loop there is a control module that decides, conditioned on the environment's representation and its own internal state, which action to take at each time step. These actions are produced by operators. Operators have a uniform interface: they admit an environment representation as input and they output a representation as well. They can therefore alter the environment and they will be used by the control module to do so until the environment reaches a target state, which represents the problem's solution. As seen in figure \ref{fig:operator}, each operator is composed of a selective input submodule, a functional submodule and a selective update submodule. Both selective submodules act as an attention mechanism and help to decouple the functionality of the operation from its interface, minimizing as well the number of parameters that a neural functional submodule would need to consume the input.

There is no imposed restriction regarding module implementations and therefore the architecture allows the building of hybrid systems. This has important consequences concerning maintenance and knowledge embedding, understood as the reutilization of existing software, manual coding or supervised training of modules. There is also the possibility of progressively evolving such a system through the replacement or addition of modules. In the latter case, the control module must be updated.

    \subsection{Motivations and architecture breakdown}
    
    The architecture we propose is mainly motivated by the idea that every problem can be decomposed in several subproblems and thus a solution to a problem's instance may be decomposed into a set of primitive operations. The observation, that an agent can decide which action to take based on its perceptions and by these means reach a certain goal, inspired us to think about problem solving in these terms. We were also aimed to increase the degree of maintainability and interchangeability of modules, thus reducing the coupling was an important matter.
    
    In the following, we introduce the main components of the architecture and describe their role in the system:
    
    \begin{itemize}
    
        \item \textit{The environment}. This represents the state of the problem and contains all information involved in the decision making process. The environment is rarely fully available to the agent, so the agent can only perceive a partial observation of it.
            
            \begin{itemize}
                \item \textit{The environment representation}. This is an abstract representation of the environment, which is sufficient to estimate the state of the problem and take the optimal action. In certain cases, where the nature of the problem is abstract enough, this environment representation is equivalent to the environment itself.

                \item \textit{The interface}. Its role is to keep the environment and its representation synchronized. Any changes that happen in the environment will be reflected in its representation and vice versa.
            \end{itemize}
            
        \item \textit{The control module}. This is the decision maker. It selects which operation should be executed, according to the current observation of the environment. This module may be equated with the agent itself, the operation modules being the tools that it uses for achieving its purpose.
        
            \begin{itemize}
                \item \textit{The digestor or perception module}. This module takes an environment representation as input, which may have unbounded size and data type, and generates a fixed size embedding from it. This module acts therefore as a feature extractor for the policy.
                
                \item \textit{The policy}. This module decides which operation to execute, conditioned on the fixed size embedding that the digestor generates.
            \end{itemize}
            
        \item \textit{Operation modules}. They implement primitive operations which will be used to solve the problem. Their architecture focuses on isolating functionality, while at the same time allowing interfacing with the environment representation.
        
            \begin{itemize}
                \item \textit{Selective input submodule}. This module filters the environment representation to select only the information relevant to the operation.
                
                \item \textit{Functional submodule}. This implements the operation's functionality.
                
                \item \textit{Selective update submodule}. This uses the output of the functional submodule to update the environment representation.
            \end{itemize}
            
    \end{itemize}
    
    Among the defined components, only the control module is strongly dependent on the problem, as it implements the logic to solve it. Other modules, like the functional and selective submodules, might well be reused to solve other problems, even from distinct problem families. The perception module, on the other hand, has a strong relationship with the environment representation, so it could mostly be reused for problems from the same family.
    
    An important appreciation is that the described architecture can also be recursive. An operation may be internally built following the same pattern and, the other way round, a modular system may be wrapped into an operation of a higher complexity. In such cases, an equivalency between the environment's API and the selective modules is implicitly established. We believe that this feature is a fundamental advantage of our modular approach, as it could be the starting point for a systematic methodology for building complex intelligent systems.
    
    \subsection{Implications regarding knowledge embedding}
    
    In this paper, we focus on studying the effects of following the modular approach in NNs. However, the proposed architecture is agnostic to the module's implementations. This gives rise to a handful of scenarios in which NNs and other implementations cooperate within the same system. The low coupling among the different modules allows embedding knowledge through manual implementation of modules, adaptation of already existing software or supervised training of NN modules.
    
    We present below some points that we believe are strong advantages:
    
    \begin{itemize}
        \item Selective submodules restrict the information that reaches the functional submodules and maintain the low coupling criteria. As their function is merely attentive, they are very eligible to be implemented manually or by reusing existing software elements.
        
        \item Interfaces of functional modules should be, by definition, compatible among different implementations. That implies that improvements in the module can come about progressively in a transparent way. Some operations would nowadays be implemented by NNs but, if someday a more reliable and efficient algorithm is discovered, they could be replaced without any negative side effect.
        
        \item The isolation of functionality in modules allows an easier analysis of neural modules, enabling the design of more efficient and explainable symbolic implementations, when applicable.
    
        \item If high-level policy sketches \cite{PolicySketches} are available, modules can learn-by-role \cite{NeuralModuleNetworks} and afterwards be analyzed. In contrast, when a policy is not available, a neural policy can be learned using RL algorithms.
    \end{itemize}

\section{Related work}

NNs have traditionally been regarded as a modular system \cite{Survey99}. At the hardware level, the computation of neurons and layers can be decomposed down to a graph of multiplications and additions. This has been exploited by GPU computation, enabling the execution of non-dependant operations in parallel, and by the development of frameworks for this kind of computation. Regarding computational motivations, the avoidance of coupling among neurons and the quest for generalization and speed of learning have been the main arguments used in favor of modularity \cite{ModularizeForGeneralization}.

The main application of modularity has been the construction of NN ensembles though, focusing on learning algorithms that automate their formation \cite{Survey99} \cite{Survey2015}. A type of ensemble with some similarities to our proposal is the Mixture of Experts \cite{MoEAdaptive} \cite{MoeHierarchical}, in which a gating network selects the output from multiple expert networks. Constructive Modularization Learning and boosting methods pursue the divide-and-conquer idea as well, although they do it through an automatic partitioning of the space. This automatic treatment of the modularization process what makes difficult to embed any kind of expertise in the system.

In \cite{NeuralModuleNetworks}, a visual question answering problem is solved with a modular NN. Each module is targeted to learn a certain operation and a module layout is dynamically generated after parsing the input question. The modules then converge to the expected functionality due to their role in such layout. This is an important step towards NN modularity, despite the modules being trained jointly.

Many of the ideas presented here have already been discussed in \cite{End2EndModuleNetworks}. They use Reinforcement Learning (RL) for training the policy module and backpropagation for the rest of the modules. However, they predict the sequence of actions in one shot and do not yet consider the possibility of implementing the feedback loop. They implicitly exploit modularity to some extent, as they pretrain the policy from expert traces and use a pretrained VGG-16 network \cite{VGG16}, but the modules are trained jointly afterwards. In \cite{ExplainableNMN} they extend this concept integrating a feedback loop, but substituting the hard attention mechanism by a soft one in order to do an end-to-end training. Thus, the modular structure is present, but the independent training is not exploited.

The idea of a NN being an agent interacting with some environment is not new and is in fact the context in which RL is defined \cite{RLIntro}. RL problems focus on learning a policy that the agent should follow in order to maximize an expected reward. In such cases there is usually no point in training operation modules, as the agent interacts simply by selecting an existing operation. RL methods would therefore be a good option to train the control module.

Our architecture proposal for the implementation of the control module has been greatly inspired by the work in Neural Programmer-Interpreters \cite{NPI}. We were also aware of the subsequent work on generalization via recursion \cite{GeneralizeViaRecursion}, but we thought it would be of greater interest to isolate the generalization effects of modularity. An important background for this work is, in fact, everything related to Neural Program Synthesis and Neural Program Induction, which are often applied to explore Domain-Specific Language (DSL) spaces. In this regard, we were inspired by concepts and techniques used in \cite{NPSwithGrammar}, \cite{NPSwithPriorityQueue} and \cite{NeuralProgramMetaInduction}.

A sort of modularity is explored in \cite{SyntheticGradients} with the main intention of decoupling the learning of the distinct layers. Although the learning has to be performed jointly, the layers can be trained asynchronously thanks to the synthetic gradient loosening the strong dependencies among them. There are also other methods that are not usually acknowledged as such but can be regarded as modular approaches to NN training. Transfer learning \cite{TransferLearning} is a common practise among the deep learning practitioners and pretrained networks are also used as feature extractors. In the field of Natural Language Processing, a word embedding module is often applied to the input one-hot vectors to transform the input space to a more efficient representation \cite{word2vec}. This module is commonly provided as a set of pretrained weights or embeddings \cite{WordTranslation}.

\section{List sorting}

In order to test the modular concept, we selected a candidate problem that complied with following desiderata:

\begin{enumerate}
    \item It has to be as simple as possible, to avoid missing the focus in modularity.
    \item It has to be feasible to solve the problem using just a small set of canonical operations.
    \item The problem complexity has to be easily identifiable and configurable.
    \item The experiments should shed light into an actual complexity-related issue.
\end{enumerate}

We found that the integer list sorting problem complied with these points. We did not worry too much about the usefulness of the problem solving itself, but rather about its simplicity and the availability of training data and execution traces. We define the list domain as integer lists containing digits from 0 to 9 and we take the \textit{Selection Sort} algorithm as reference, which has $\mathcal{O}(n^2)$ time complexity, but is simple in its definition. This implies that the environment will be comprised by a list of integers and two pointers, which we name A and B. We associate the complexity level to the maximal training lists length, because it will determine the internal recurrence level of the modules. So, a network trained on complexity $N$ is expected to sort lists with up to $N$ digits. This setup also leads us to the following canonical operations:

\begin{itemize}
    \item \texttt{mova}. Moves the pointer A one position to the right.
    \item \texttt{movb}. Moves the pointer B one position to the right.
    \item \texttt{retb}. Returns the pointer B to the position to the right of the pointer A.
    \item \texttt{swap}. Exchanges the values located at the positions pointed by A and B.
    \item \texttt{EOP}. Leaves the representation unchanged and marks the end of execution of the perception-action loop.
\end{itemize}

Each problem instance can be solved based on this set of primitive operations. At the beginning, the agent starts with a zeroed internal state and the environment in an initial state, where the pointers A and B are pointing to the first and second digits respectively. We say that the environment state is final if both pointers are pointing to the last digit of the list. The execution nevertheless stops when the agent selects the \texttt{EOP} operator. The goal is to use a NN to solve the proposed sorting problem, after training it in a supervised fashion.

Because we deal with sequential data, the most straightforward choice was to use Recurrent Neural Networks (RNN) to implement the different submodules. This brings up the common issues related to RNNs, which are for the most part the gradient vanishing and the parallelization difficulties due to sequential dependencies. Our experiments will offer some clues about how a modular approach can help to cope with such issues.

\section{Dynamical data generation}

Our dataset comprises all possible sequences with digits from $0$ to $9$, with lengths that go from 2 digits long up to a length of $N$. We represent those digits with one-hot vectors and we pad with zero-vectors the positions coming after the list has reached its end.

For training, we randomly generate lists with lengths within the range $[2,N]$. For testing, we evaluate the network on samples of the maximum training length $N$. We also generate algorithmic traces. After lists are sampled, we generate the traces based on the \textit{Selection Sort} algorithm, producing the operations applied at each time step as well as the intermediate results. These traces are intended to emulate the availability of expert knowledge in the form of execution examples, just as if they were previously recorded, and allow us to measure the correctness of execution.

\begin{figure}
    \centering
    \includegraphics[keepaspectratio,width=\columnwidth]{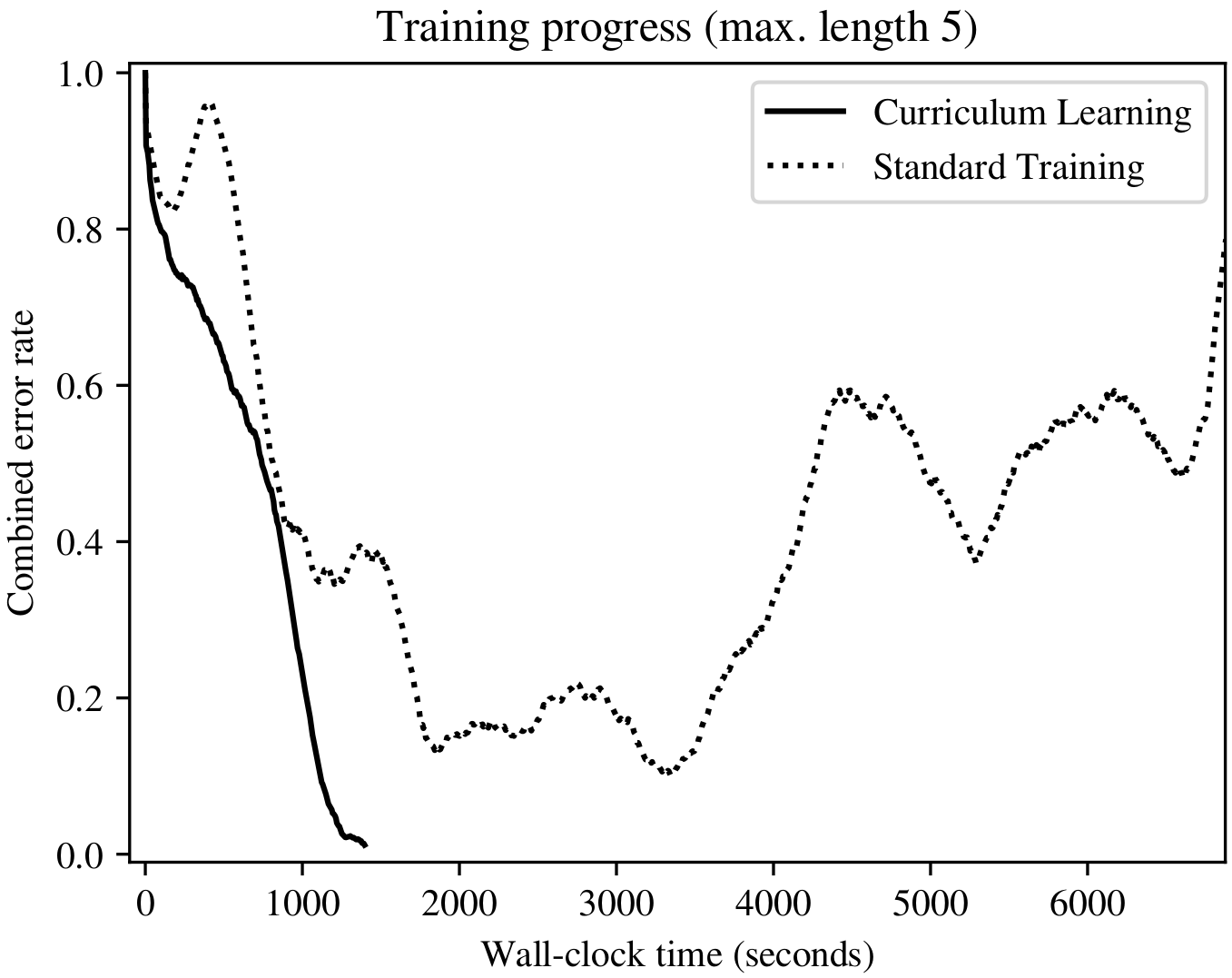}
    \caption{Progress of the error rate for standard training and curriculum learning. Standard training was stopped before quantitative convergence, after 7 hours of execution without improving.}
    \label{fig:curriculum_learning}
\end{figure}

We draw list lengths from a beta distribution that depends on the training accuracy $B(\alpha,\beta)$, where $\alpha = 1 + \text{accuracy}$, $\beta = 1 + (1 - \text{accuracy})$ and $\text{accuracy} \in [0,1]$. In this way, we can start training on shorter lengths and by the end of the training we mainly sample long lists. We found this sampling method to be very advantageous with respect to the uniform sampling. In figure \ref{fig:curriculum_learning} we show a comparison between both sampling methods. There, we can see the inability of the model to converge under uniformly sampled batches, seemingly because of the complexity residing in the longer sequences and their infrequent appearance under such sampling conditions.

\section{Neural Network architecture}

We intend to evaluate the impact of modularization and modular training in a NN and assess its technical implications, transferring the proposed abstract architecture to a particular case. Therefore, we have implemented a modular NN architecture, in which every module (except selection submodules) is a NN.

This layout enables us to train the network's modules independently and assemble them afterwards, as well as to treat the network as a whole and train it end to end, in the most common fashion. In this way we are able to make a fair comparison between the modular and the monolithic approach. In all cases we will train the NN in a supervised manner, based on (input, output) example pairs.

\begin{figure}
    \centering
    \includegraphics[keepaspectratio,width=\columnwidth]{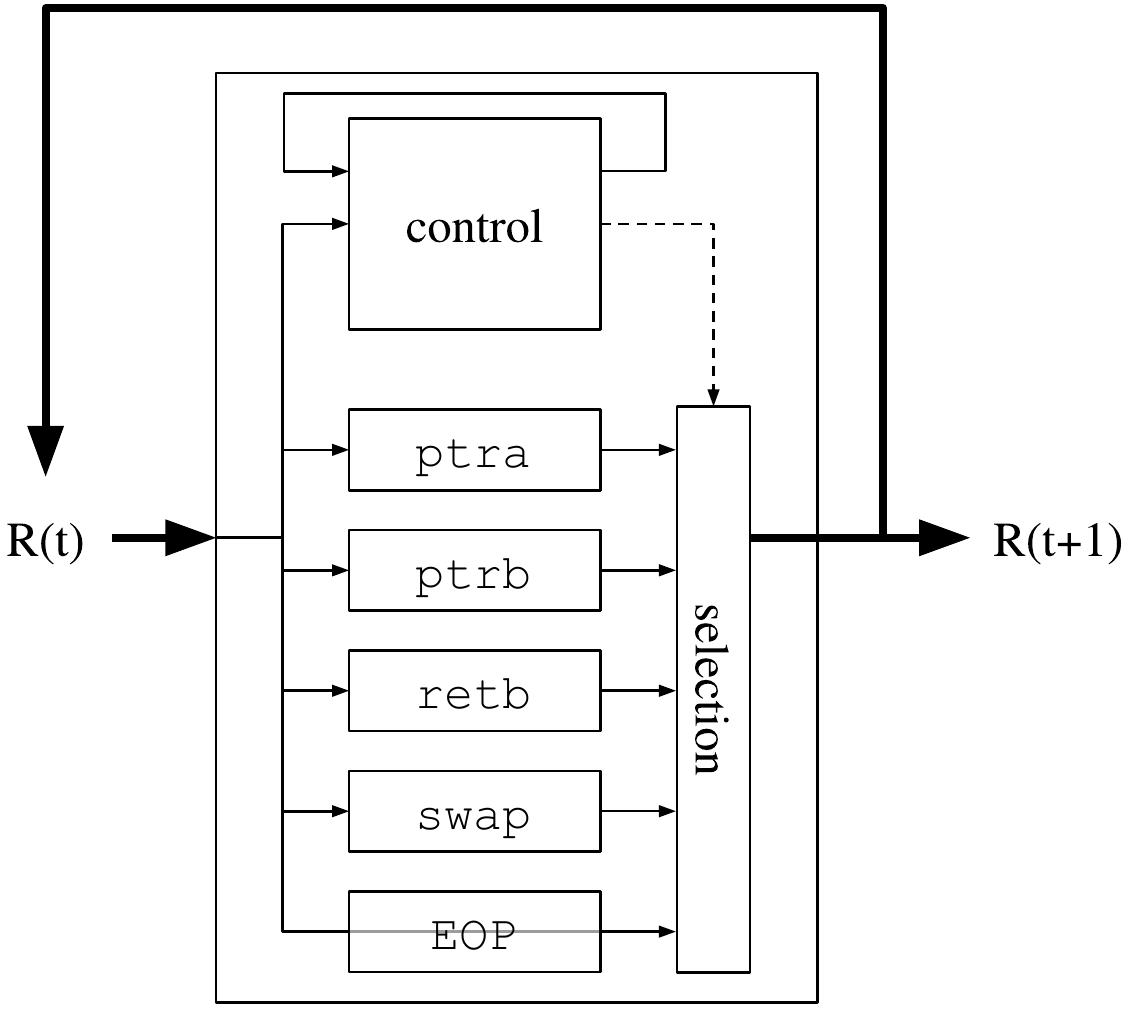}
    \caption{Implementation of the perception-action loop for the list sorting problem. At each time step, the control module selects the output of one operation, which substitutes the previous representation.}
    \label{fig:main_loop_impl}
\end{figure}

The network can interact with a predefined environment by perceiving it and acting upon it. At each execution step, the current state representation of the environment is fed to all existing operations. Then, the control module, conditioned on the current and past states, selects which operation will be run, its output becoming the next representation (figure \ref{fig:main_loop_impl}). In our implementation, we omit the interface against the environment and establish an equivalency between the environment and its representation.

As said before, the environment has three elements: the list and two pointers (A and B). The list is a sequence of one-hot vectors, encoding integer digits that go from 0 to 9. Null values are encoded by a zero vector. The pointers are sequences of values comprised in the range $[0,1]$, which indicate the presence of the pointer at each position. A value over $0.5$ means that the pointer is present at that position.

Each operation module is implemented following the main modular concept (figure \ref{fig:operator}). We only allow the functional submodules to be implemented by a NN and build the selective submodules programmatically. In \texttt{ptra} and \texttt{ptrb}, the same pointer is selected as input and output. \texttt{retb} selects A as input and updates B. \texttt{swap} merges the list and both pointers into a single tensor to build the input and updates only the list.

\begin{figure}
    \centering
    \includegraphics[keepaspectratio,width=\columnwidth]{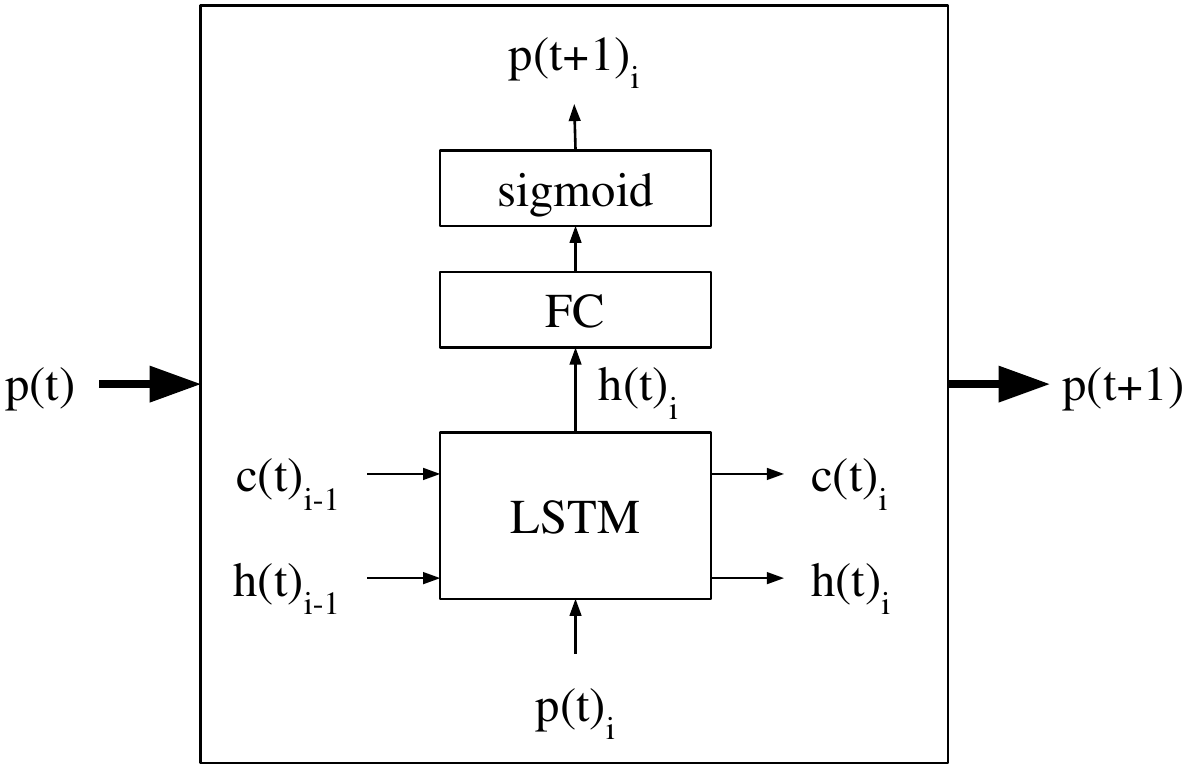}
    \caption{Architecture of the pointers' functional submodule, with an LSTM and a fully connected output layer with sigmoid activation. $c(t)_i$ and $h(t)_i$ are the LSTM's internal state and output at each time step and position $i$.}
    \label{fig:pointer_cell}
\end{figure}

\begin{figure}
    \centering
    \includegraphics[keepaspectratio,scale=0.7]{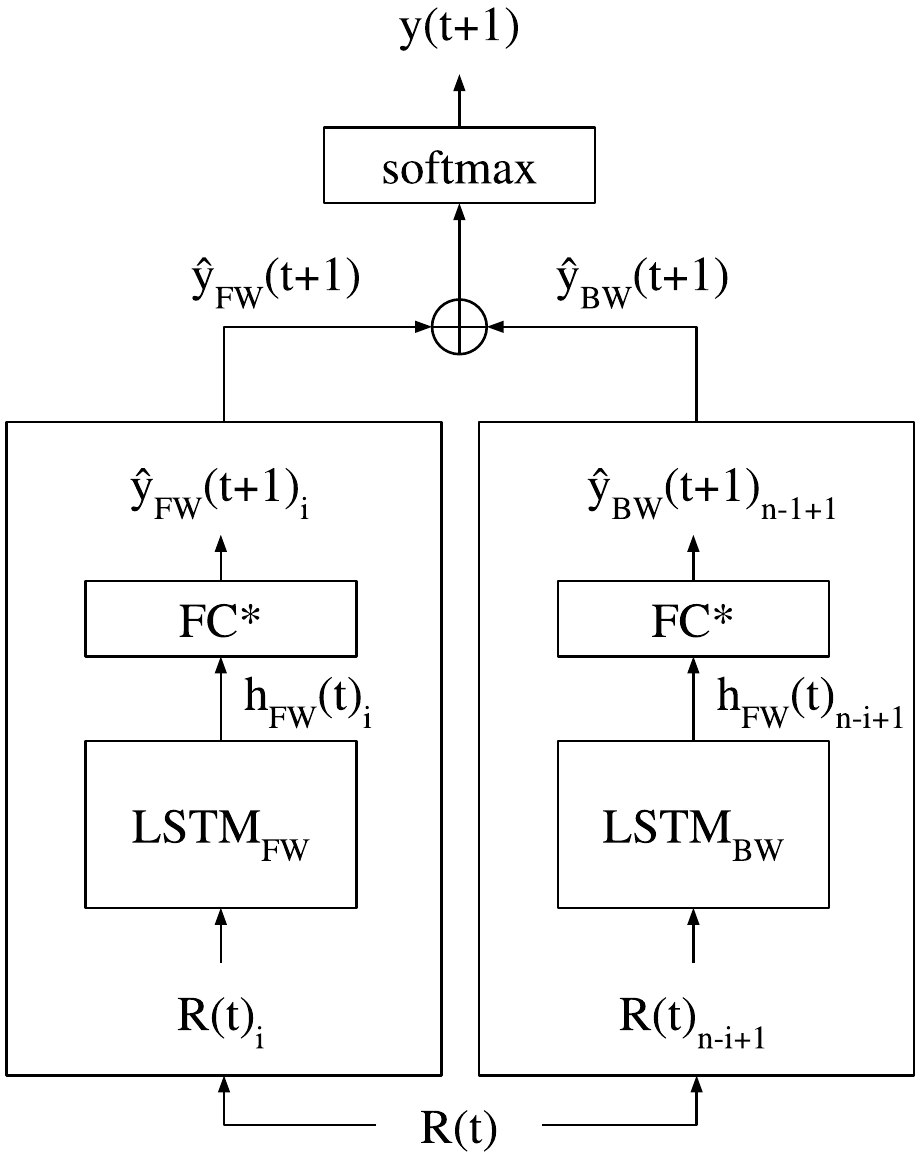}
    \caption{Architecture of the \texttt{swap} functional module. The entire representation is merged into a one single tensor and fed to a bidirectional LSTM. The outputs pass through a fully connected layer and are then merged by addition. *Fully connected layers share parameters.}
    \label{fig:swap_cell}
\end{figure}

The architecture of each functional submodule is different, depending on the nature of the operation, but they all use at least one LSTM cell, which always has 100 units. Pointer operations use an LSTM cell, followed by a fully connected layer with a single output and a sigmoid activation (figure \ref{fig:pointer_cell}). The swap submodule is based on a bidirectional LSTM. The output sequences from both forwards and backwards LSTM are fed to the same fully connected layer with 11 outputs and summed afterwards. This resulting sequence is then passed through a softmax activation (figure \ref{fig:swap_cell}). We discard the eleventh element of the softmax output to enable the generation of zero vectors. The \texttt{EOP} operation does not follow the general operator architecture, as it just forwards the input representation.

\begin{figure}
    \centering
    \includegraphics[keepaspectratio,scale=0.7]{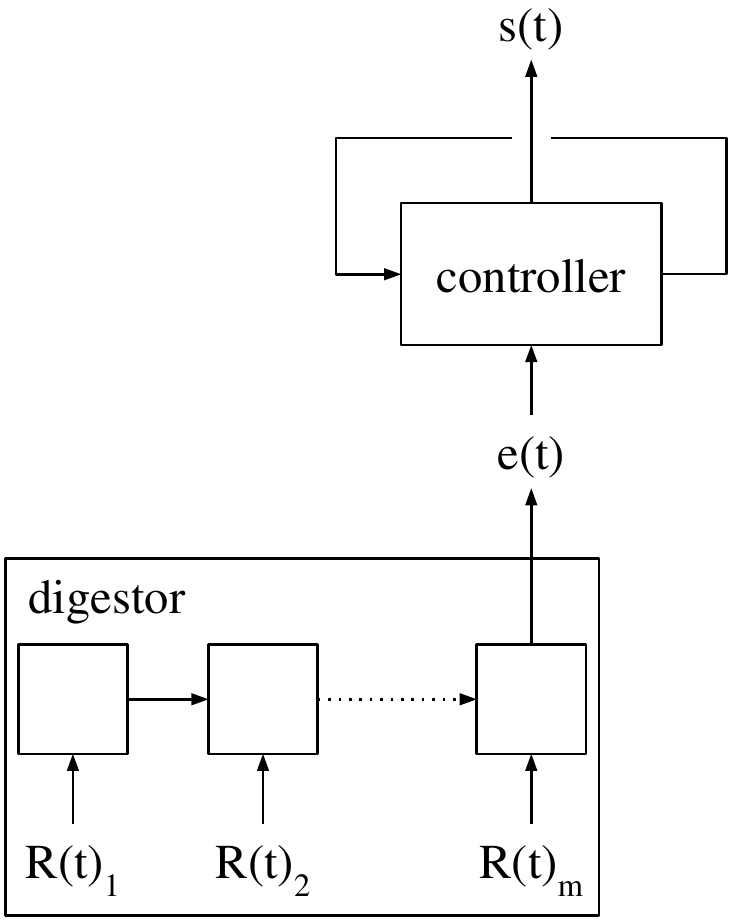}
    \caption{The \textit{digestor} creates a fixed-size embedding $e(t)$ from the state representation and the \textit{controller} takes it as input at every execution step. Conditioned on the embedding and its past state, it outputs the selection vector $s(t)$.}
    \label{fig:control_cell}
\end{figure}

The control module is intended to be capable of: 1) perceiving the environment's state representation, regardless of its length and 2) conditioning itself on previous states and actions. Therefore, its architecture is based on two LSTM cells, running at two different recurrence levels (figure \ref{fig:control_cell}). The first LSTM, which we call the \textit{digestor}, consumes the state representation one position at a time and produces a fixed-size embedding at the last position. This fixed-size embedding is fed to the second LSTM (the controlling policy) as input. While the \textit{digestor}'s internal state gets zeroed before consuming each state, the \textit{controller} ticks at a lower rate and gets to keep its internal state during the whole sorting sequence.

\section{Experimental setup}

Each agent configuration is trained until specific stop criteria are fulfilled. Our intention is that training conditions for all configurations are as equal as possible. We make use of two distinct measures for determining when a model has reached a satisfactory training state and we keep a moving average for each one of them.

\begin{itemize}
    \item The \textit{quantitative} measure focuses on functionality and depends on the corresponding error rates. An output list is counted as an error when there is any mismatch with respect to the expected one. A pointer is considered valid if the only value above $0.5$ corresponds to the right position. Otherwise it is taken as an error.
    \item The \textit{qualitative} measure depends on the percentage of output values that do not comply with a certain saturation requirement, which is that the difference with respect to the one-hot label is not greater than $0.1$.
\end{itemize}

The monolithic configuration is trained until the quantitative measure reaches values below 1\%. The training of the operation modules takes also the qualitative measure into account and only stops when both measures are below 1\%. As a measure to constrain the training time, the training of the monolithic configuration is also stopped if the progress of the loss value becomes stagnant or if the loss falls below \texttt{1e-6}.

It is convenient that the modules work well when several operations are concatenated and that is why we require the quality criterion and why we train them under noisy conditions. In this regard, we apply to the inputs noise sampled from a uniform distribution to deviate them from pure $\{0,1\}$ values up to a $0.4$ difference (eq. \ref{eq:uniform-noise}). Inputs that represent a one-hot vector are extended to 11 elements before adding the uniform noise and passing them through a softmax-like function (eq. \ref{eq:softmax-noise}) in order to keep the values on the softmax manifold. The eleventh element is then discarded again to allow the generation of zero vectors.

\begin{equation}
    \hat{x}_{uniform} = | x - U(0, 0.4) |
    \label{eq:uniform-noise}
\end{equation}

\begin{equation}
    \hat{x}_{softmax} = \text{softmax}(\hat{x}_{uniform} \cdot 100)
    \label{eq:softmax-noise}
\end{equation}

Every configuration is trained in a supervised fashion, making use of the cross-entropy loss and Adam \cite{Adam} as the optimizer. The learning rate is kept the same (\texttt{1e-3}) across all configurations. The cross-entropy loss is generally computed with respect to the corresponding output, with the exception of the monolithic configuration, which is provided with an additional cross-entropy loss computed over the selection vectors. This last bit is relevant because it enables the monolithic configuration to learn the algorithm and the operations simultaneously.

We have built three different training setups: module-wise training, monolithic training and staged training. In the staged training, each module is trained independently, but after every 100 training iterations the modules are assembled and tested in the assembled configuration.

The monolithic configuration is the most unstable and the results vary significantly between different runs. Thus we average the results obtained through 5 runs in order to alleviate this effect. To ease training under this configuration, we also make the selection provided by the training trace override the control output, as it is done in \cite{NPSwithGrammar}. This mechanism is deactivated during testing, but during training allows the operations to converge faster, regardless of the performance of the control module. 

We tried to compare results against a baseline model, using a multilayer bidirectional LSTM, but such configuration did not seem to converge to any valid solution, so we decided to discard it for the sake of clarity.

\section{Experimental results}

\begin{figure}
    \centering
    \includegraphics[keepaspectratio,width=\columnwidth]{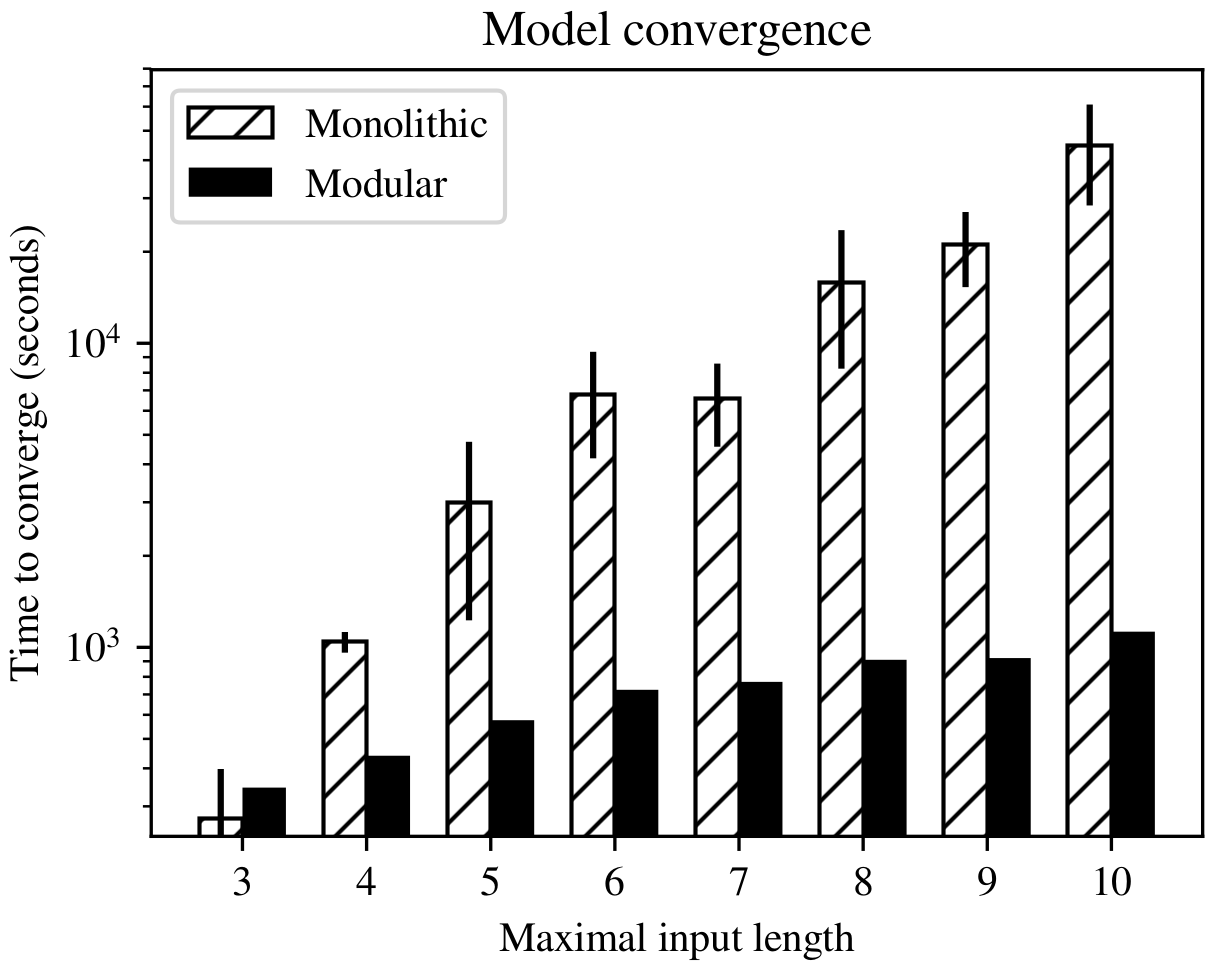}
    \caption{Convergence times for modular and monolithic configurations. Training times do not only get longer with longer input sequences, but also become more unstable. We hypothesize this is because of the high recurrence. The time scale is logarithmic.}
    \label{fig:convergence}
\end{figure}

After training both modular and monolithic configurations, we saw that the training time is orders of magnitude shorter when it is trained modular-wise (figure \ref{fig:convergence}), despite the requirements being stronger. It is important to stress that in this case we considered only the worst case scenario, so we count the time that takes to train the modules sequentially. The training time can be reduced even further if all modules are trained in parallel.

\begin{figure}
    \centering
    \includegraphics[keepaspectratio,width=\columnwidth]{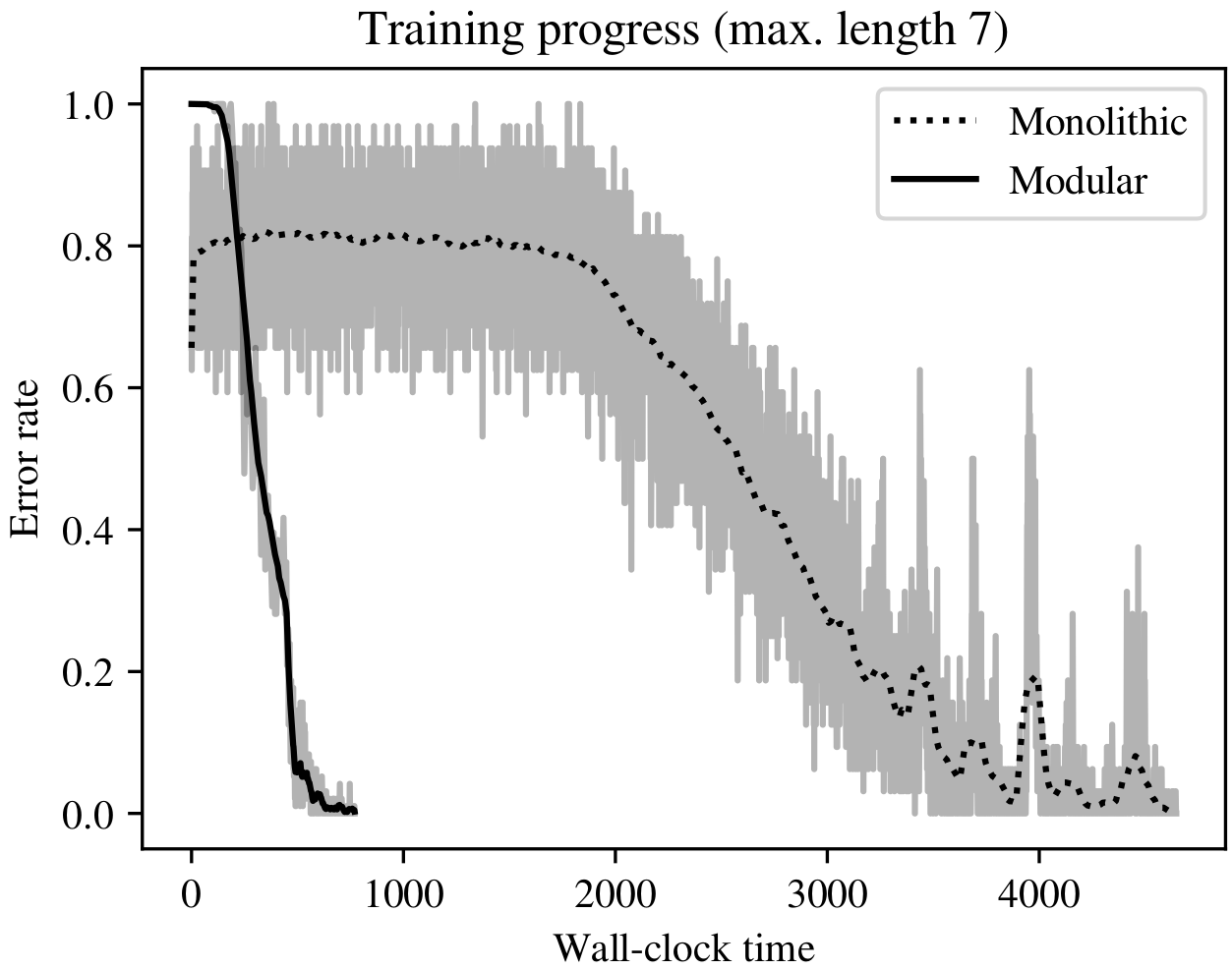}
    \includegraphics[keepaspectratio,width=\columnwidth]{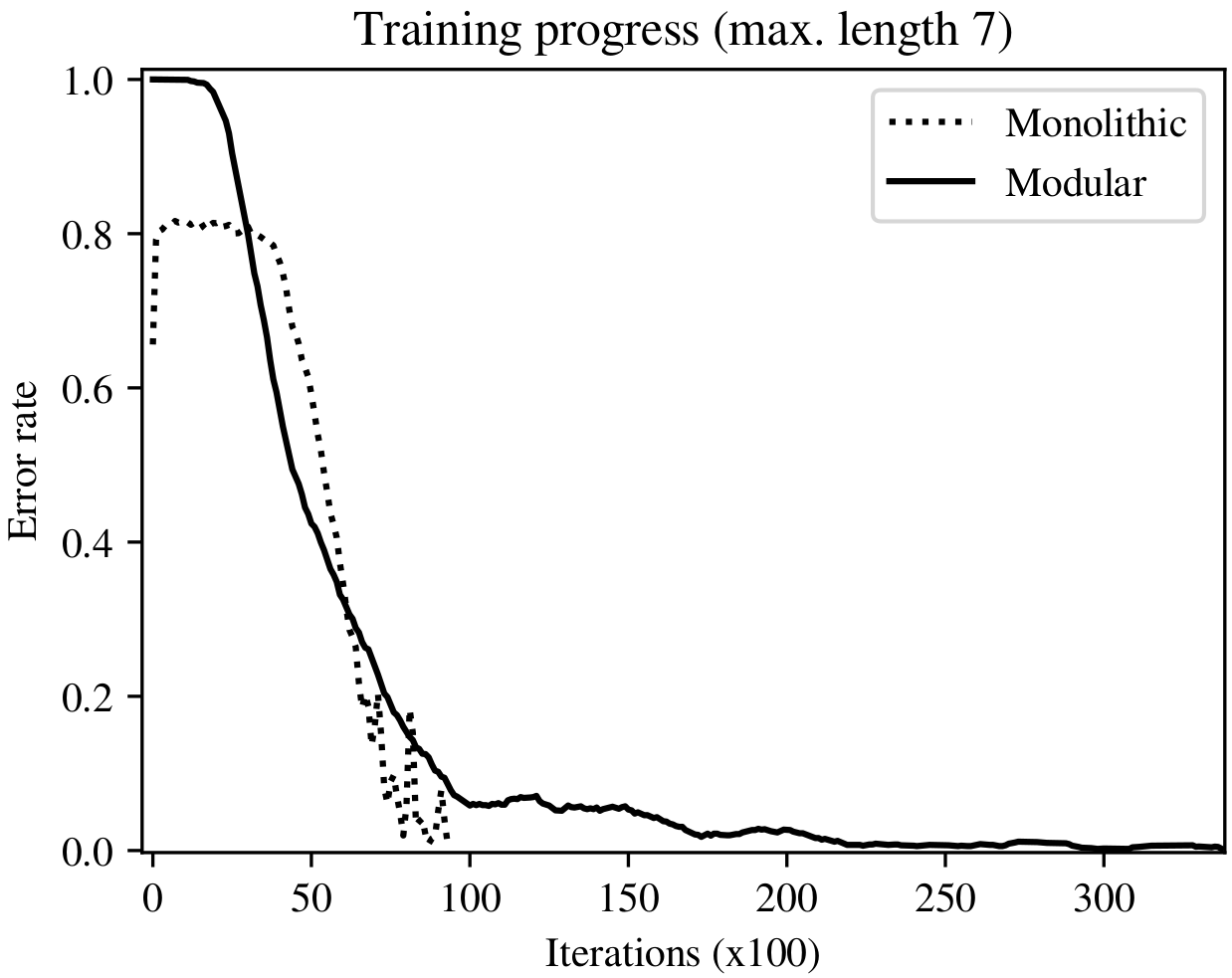}
    \caption{Error rate curves during training for modular and monolithic configurations, with respect to time (top) and training iterations (bottom).}
    \label{fig:staged_curves}
\end{figure}

\begin{figure}
    \centering
    \includegraphics[keepaspectratio,width=\columnwidth]{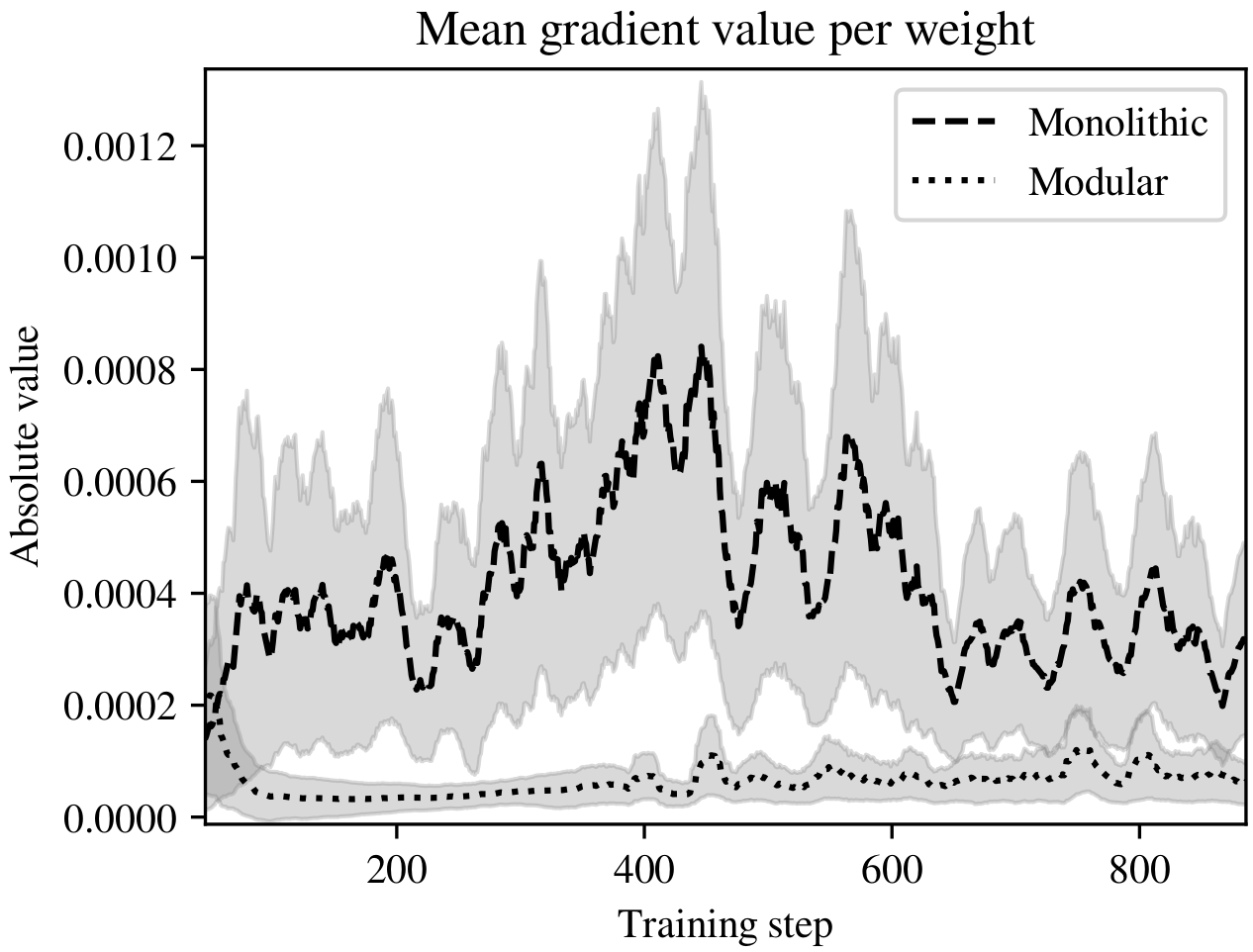}
    \caption{Mean absolute value of the gradient at the weights for each configuration. This data was obtained while training with lists of 7 digits maximum. Obtained data was slightly smoothed to help visualization. The grey shadow represents the standard deviation.}
    \label{fig:gradient}
\end{figure}

We also see in figure \ref{fig:staged_curves} how each training progresses in a very different manner, the modular configuration needing much less time to reach low error rates than the monolithic one. Though it takes more training iterations, they are faster to compute and the error rate is more stable. We were curious about this behaviour and we conducted additional measurements regarding the gradient (figure \ref{fig:gradient}). We then saw then that the gradient is much richer in the monolithic case, with a higher mean absolute value per parameter and greater variations. This makes sense, as the back propagation through time accumulates gradient at every time step and the monolithic configuration has a recurrence of $O(N^2)$, so it is more informative and can capture complex relations, even between modules.

\begin{figure}
    \centering
    \includegraphics[keepaspectratio,width=\columnwidth]{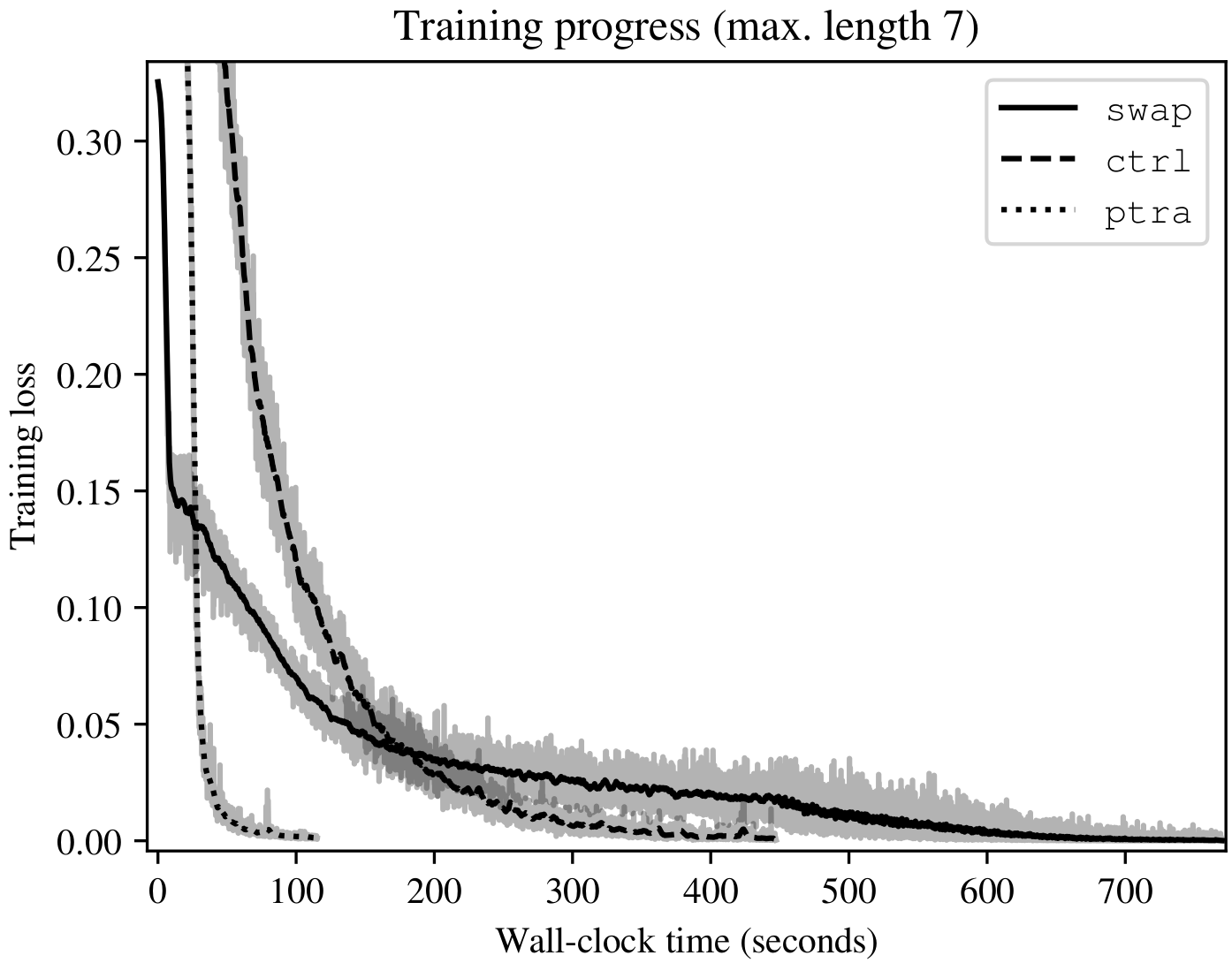}
    \caption{Progress of the training loss for the different operations during training with lists of maximal length 7. We only show \texttt{ptra} because the complexity is the same as in \texttt{ptrb} and \texttt{retb}.}
    \label{fig:progress_modules}
\end{figure}

By observing the training data more thoroughly, we can appreciate the relative complexity of the different modules. In figure \ref{fig:progress_modules} we plot the loss curves for each module in the network when trained independently. Pointer operations converge very quickly, as they only learn to delay the input in one time step. The swap operation needs more time, but thanks to the bidirectional configuration each LSTM just needs to remember one digit (listing \ref{text:example_swap}). Surprisingly, the control module does not need as much time as the swap module to converge, even when having to learn how to digest the list into an embedded representation and to use its internal memory for remembering past actions. This could again be a consequence of a richer gradient.

\begin{lstlisting}[language=C,
    caption={Example of a bidirectional LSTM performing the \texttt{swap} operation onto a list. An underscore represents a zero-vector.},
    label={text:example_swap}]
L = 3,9,5,4,_
A = 0,1,0,0,0
B = 0,0,0,1,0

Forward  LSTM Output = 3,_,5,9,_
Backward LSTM Output = _,4,_,_,_
Merge by addition ---> 3,4,5,9,_
\end{lstlisting}

\begin{figure}
    \centering
    \includegraphics[keepaspectratio,width=\columnwidth]{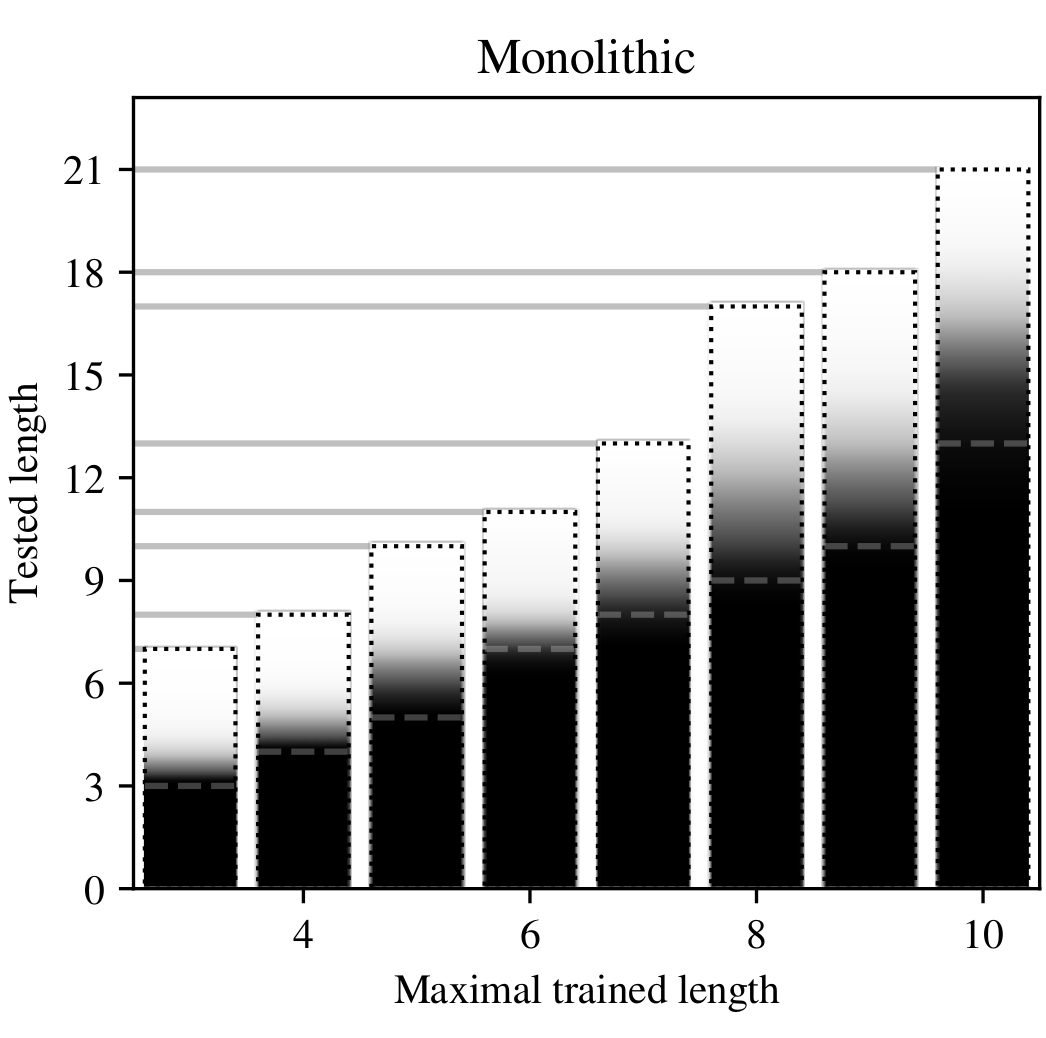}
    \includegraphics[keepaspectratio,width=\columnwidth]{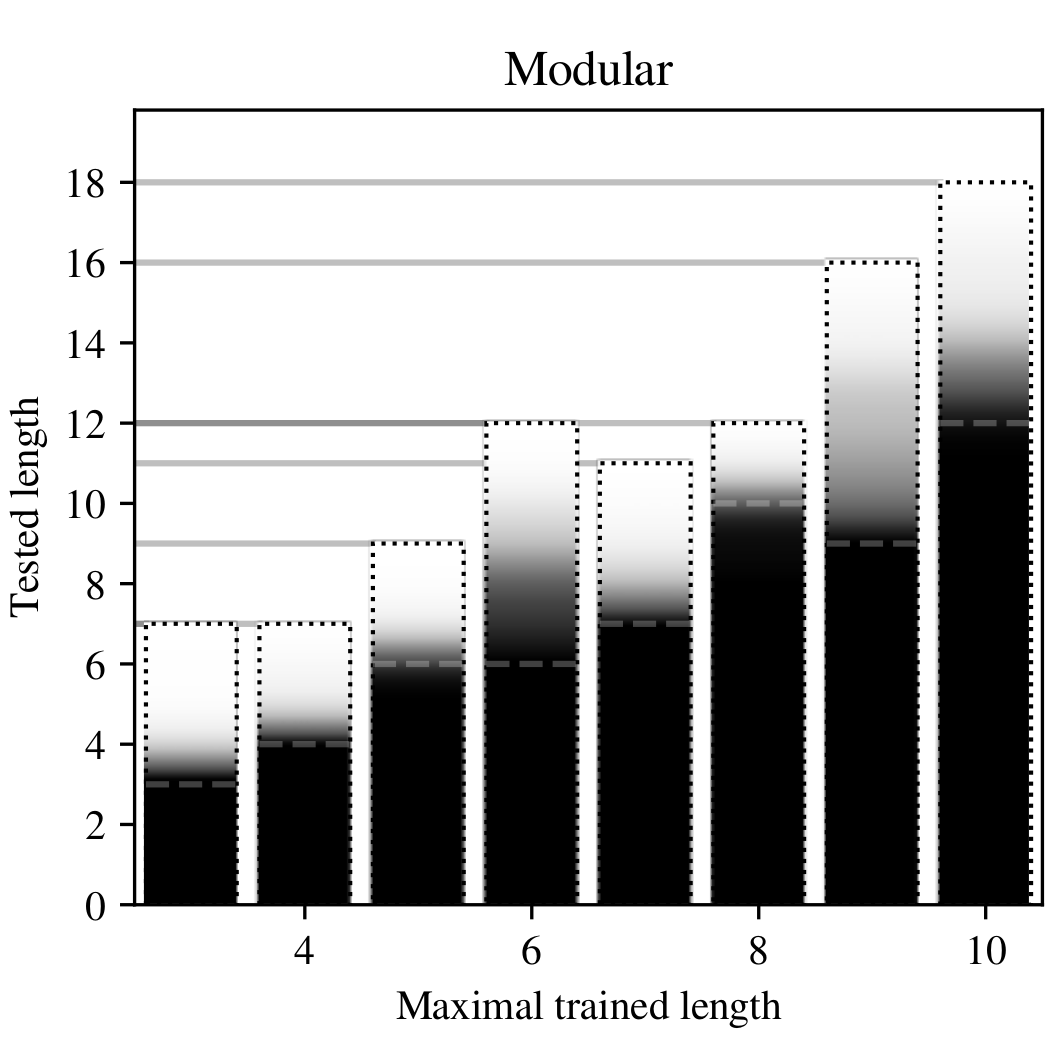}
    \includegraphics[keepaspectratio,width=\columnwidth]{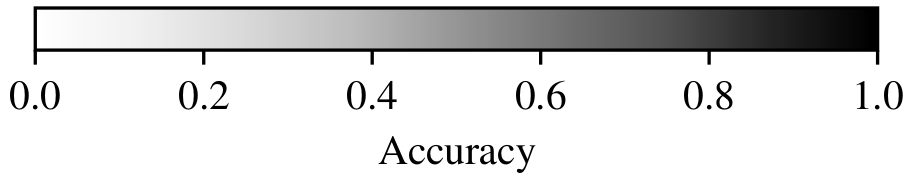}
    \caption{Generalization tests for monolithic (top) and modular (bottom) configurations. Horizontal lines mark the length where $0$ accuracy is achieved. A dashed line points where the accuracy passes $0.9$.}
    \label{fig:generalization}
\end{figure}

Regarding generalization, in figure \ref{fig:generalization} we show the behaviour of each configuration when tested on list lengths not seen during training. The monolithic configuration generalizes better to longer lists and its performance degrades slowly and smoothly than the modular one. This seems to contradict the common belief that modular NNs are able to achieve better generalization. However, we hypothesized this could happen due to the additional restrictions applied to the modular training, such as the random input noise and the output saturation requirements.

\begin{figure}
    \centering
    \includegraphics[keepaspectratio,width=\columnwidth]{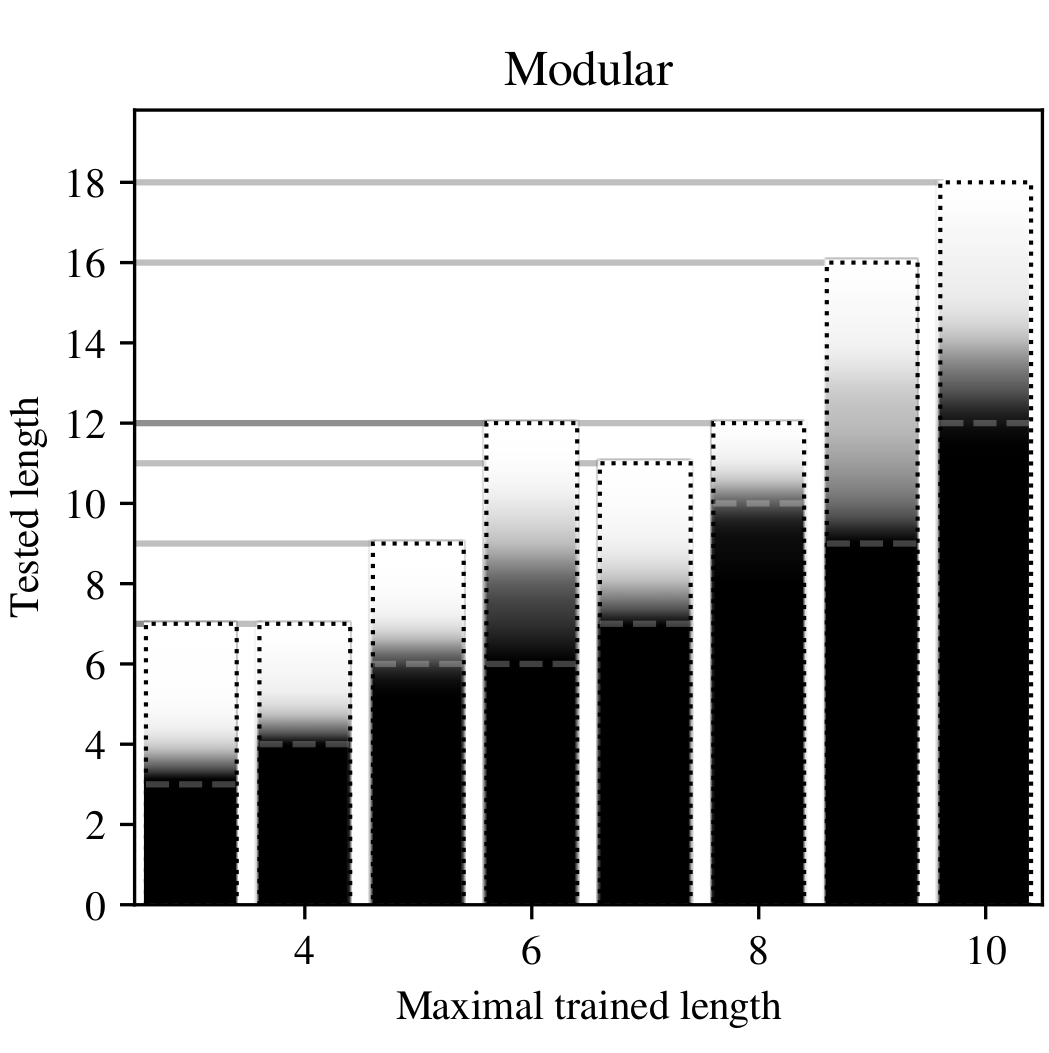}
    \caption{Generalization tests for the modular configuration after being trained with the corrected learning rate.}
    \label{fig:generalization_extra}
\end{figure}

In the modular configuration, we tried to compensate the lack of such gradient quality with ad-hoc loss functions and training conditions, but adding such priors can also backlash. This is therefore a phenomenon that should be considered when designing modular NNs. In this case, we tried a learning rate 5 times higher to compensate for the lower gradient and we experienced a training time reduction of more than a half, with a slight increase in generalization (figure \ref{fig:generalization_extra}). Further study of the special treatment for modular NNs could be part of future research.

\section{Conclusions}

We proposed a modular approach to NNs design based on a perception-action loop, in which the whole system is functionally divided in several modules with standardized interfaces. These modules are all liable to be trained, either independently or jointly, or also explicitly specified by a human programmer. We have shown how a list sorting problem can be solved by a NN following this modular architecture and how modularity has a very positive impact on training speed and stability. There seems to be a trade-off with respect to the generalization and number of training steps though, which somehow suffer from not having access to a global gradient and excessive restrictions during training. We give insights in this phenomena and suggestions to address them.

Designing modular NNs can lead to a better utilization of computational resources and data available, as well as an easier integration of expert knowledge, as it is discussed in the document. NN modules under this architecture are easily upgradeable or interchangeable by alternative implementations. Future research should explore this kind of scenarios and practical implementations of the modular concept. The effects of modular training on non-recurrent NN modules should be studied as well. Moreover, what we have introduced is an initial approach, so further investigations may reveal a variety of faults and improvements, in particular regarding the application of this concept to problems of higher complexity.

\section{Acknowledgements}

We would like to thank the reviewers for their valuable opinions and suggestions, which have helped us to substantially improve the quality of this article.

\bibliography{main.bib}
\bibliographystyle{aaai}

\end{document}